\documentclass[letterpaper, 10 pt, conference]{ieeeconf}  

\IEEEoverridecommandlockouts                              

\usepackage{amsmath}
\usepackage{cite}
\usepackage{amsmath,amssymb,amsfonts}
\usepackage{algorithmic}
\usepackage{graphicx}
\usepackage{textcomp}
\usepackage{hyperref}
\usepackage{xcolor}
\usepackage{subfig}
\usepackage{multirow}

\def\BibTeX{{\rm B\kern-.05em{\sc i\kern-.025em b}\kern-.08em
    T\kern-.1667em\lower.7ex\hbox{E}\kern-.125emX}}

\begin{document}

\title{\LARGE \bf
A Comparison of Modern General-Purpose Visual SLAM Approaches}

\author{
\authorblockN{Alexey Merzlyakov}
\authorblockA{
\textit{Samsung Research Russia}\\
alexey.merzlyakov@samsung.com}
\and
\authorblockN{Steve Macenski}
\authorblockA{
\textit{Samsung Research America}\\
s.macenski@samsung.com}
}

\maketitle
\thispagestyle{empty}
\pagestyle{empty}

\begin{abstract}
Advancing maturity in mobile and legged robotics technologies is changing the landscapes where robots are being deployed and found.
This innovation calls for a transformation in simultaneous localization and mapping (SLAM) systems to support this new generation of service and consumer robots. 
No longer can traditionally robust 2D lidar systems dominate while robots are being deployed in multi-story indoor, outdoor unstructured, and urban domains with increasingly inexpensive stereo and RGB-D cameras.
Visual SLAM (VSLAM) systems have been a topic of study for decades and a small number of openly available implementations have stood out: ORB-SLAM3, OpenVSLAM and RTABMap.

This paper presents a comparison of these 3 modern, feature rich, and uniquely robust VSLAM techniques that have yet to be benchmarked against each other, using several different datasets spanning multiple domains negotiated by service robots.
ORB-SLAM3 and OpenVSLAM each were not compared against at least one of these datasets previously in literature and we provide insight through this lens.
This analysis is motivated to find general purpose, feature complete, and multi-domain VSLAM options to support a broad class of robot applications for integration into the new and improved ROS 2 Nav2 System as suitable alternatives to traditional 2D lidar solutions.

\end{abstract}


\section{Introduction}
\label{sec:introduction}
Simultaneous localization and mapping (SLAM) has been a field of study for many decades.
From radars and range finders, to cameras and lasers, many modalities of SLAM have been developed to solve the fundamental problem of finding sensor poses in a global representation.
The development of 2D laser scanners, while expensive, was a game changer that unlocked the mass deployment of service robots seen today.
2D SLAM and localization techniques have led the way for reliable and computationally efficient positioning in large scale environments for much of the last decade.  
As robot costs are being driven down to enable large scale fleets, sale to consumers, and application to new problems, the cost of laser scanners has become a significant bottleneck.
Further, robots are now being deployed in multi-story construction sites, urban cities, and all-terrain areas not suitable for 2D techniques - regardless of sensor price.
Robust and mature approaches relying on cameras and comparatively low cost RGB-D sensors will be crucial for enabling this next wave of robot applications. 

Many visual SLAM (VSLAM) techniques have been proposed and studied in literature.
However, markedly fewer have been proposed with sufficient maturity to be deployed on robots in real-world environments for the long haul \cite{iros2019}.
Features such as pure localization, re-localization of a lost track, resource efficiency, loop closure, reliability, and support for a broad range of sensor types are givens in 2D SLAM, but are not all frequently found in VSLAM.
Implementations that contain these features have gained substantial popularity and are well suited for the needs of mobile robots of today and tomorrow.

In this paper, we compare 3 modern, robust, and feature rich visual SLAM techniques: ORB-SLAM3\cite{ORB-SLAM3}, OpenVSLAM\cite{OpenVSLAM}, and RTABMap\cite{RTABMap}.
The purpose of this comparison is to identify robust, multi-domain visual SLAM options which may be suitable replacements for 2D SLAM for a broad class of service robot uses.
These are compared in multiple operating domains with several sensors to showcase each technique's ability to generalize across many environments and sensor configurations. 
We also study the maturity and features each provide, enabling them to be considered for service robot use.

This comparison is additionally novel by offering a benchmark using 3 datasets and 3 recent VSLAM techniques that have not been formally compared before.
ORB-SLAM3 and OpenVSLAM are both run against at least one new dataset that has not been published in literature previously (TUM RGB-D).
The comparison is also made unique through the focus on general purpose techniques that perform well across different domains and sensor configurations.
These techniques are not only analyzed on their core algorithm performance on datasets, but also their features and reliability to assess their ability to take the place of lidar-based SLAM solutions. 

In studying this problem, we seek to accelerate the utilization of VSLAM in service robotics; enabling new application of robotics to previously priced out or non-planar applications.
The mobile robotics and service robotics fields has not previously had reliable and feature-complete "workhorse" VSLAM options to build into navigation systems for general use.
The results of this comparison are integrated into the new and improved Nav2 (ROS 2 Navigation).
Nav2 offers a parallel visual SLAM and localization integration to complement the existing 2D options to further accelerate and simplify research and industry adoption \cite{Nav2}.

\begin{table*}[ht]
 \caption{Summary of most representative Visual SLAM/Visual Odometry approaches}
 \label{tab:vslam-approaches}
 \begin{center}
 \begin{tabular}{ |c|c|c|c|c|c|c|c|c|c|c| }
  \hline
  \multirow{2}{*}{VSLAM} & \multirow{2}{*}{Type} & \multirow{2}{*}{Mono} & \multirow{2}{*}{Stereo} & \multirow{2}{*}{RGB-D} & \multirow{2}{*}{Fisheye} & \multirow{2}{*}{IMU} & Pure & \multirow{2}{*}{Re-Localization} & Loop & \multirow{2}{*}{Dev.status} \\
  & & & & & & & Localization & & Closing & \\
  \hline
  \hline
  ORB-SLAM3\cite{ORB-SLAM3} & SLAM & \checkmark & \checkmark & \checkmark & \checkmark & \checkmark & \checkmark & \checkmark & \checkmark & Sep 2020\\
  OpenVSLAM\cite{OpenVSLAM} & SLAM & \checkmark & \checkmark & \checkmark & \checkmark & - & \checkmark & \checkmark & \checkmark & Active$^1$ \\
  RTABMap\cite{RTABMap} & SLAM & - & \checkmark & \checkmark & \checkmark & - & \checkmark & \checkmark & \checkmark & Active \\
  Kimera\cite{Kimera} & VIO/SLAM & \checkmark$^2$ & \checkmark$^2$ & - & - & \checkmark & - & - & \checkmark & Active \\
  VINS-Fusion\cite{VINS-Mono, VINS-Fusion} & VIO & \checkmark$^2$ & \checkmark & - & \checkmark & \checkmark & \checkmark & \checkmark & \checkmark & Oct 2019 \\
  SVO\cite{SVO1, SVO2} & VO & \checkmark & \checkmark & - & \checkmark & - & - & \checkmark & - & May 2017 \\
  DSO\cite{DSO} & VO & \checkmark & \checkmark & - & \checkmark & - & - & - & - & Dec 2018 \\
  LSD-SLAM\cite{LSD-SLAM1} & SLAM & \checkmark & \checkmark & - & - & - & - & \checkmark & \checkmark & Dec 2014 \\
  PL-SLAM\cite{PL-SLAM} & SLAM & - & \checkmark & - & - & - & - & - & \checkmark & Nov 2018 \\
  OKVIS\cite{OKVIS1} & VIO & \checkmark$^2$ & \checkmark$^2$ & - & - & \checkmark & - & - & - & Jul 2016 \\
  \hline
  \multicolumn{11}{l}{} \\
  \multicolumn{11}{l}{$^1$ Despite OpenVSLAM source are closed at the moment of publication, significant development continues in OpenVSLAM-Community fork}\\
  \multicolumn{11}{l}{$^2$ Pure sensor mode without IMU is not supported for the method}\\
 \end{tabular}
 
 \end{center}
\end{table*}

\section{Related Work}
\label{sec:related_work}


There is a great variety of visual SLAM approaches described in literature - a list of popular and relatively modern VSLAM works is shown in Table \ref{tab:vslam-approaches}.
All of these techniques have reference implementations or binaries available to test against and compare with new datasets.

The aim of this work is to identify VSLAM options that can be deployed on many different service, legged, and mobile robotics applications reliably.
This requires minimum features and requirements that not all of the techniques in Table \ref{tab:vslam-approaches} meet.
To be considered for evaluation as a general purpose technique, the method must provide (at bare minimum) support for loop closures, re-localization, and pure localization.
This requirement is set to ensure that the compared solution can both map a space to be globally consistent and is able to localize within this map over multiple days, months, or even years.
Further, it must also support the two most common sensor configurations found on service robots to replace or augment laser scanners: stereo and RGB-D.
It is very common for indoor robots to make use of active RGB-D cameras while outdoor robots make use of passive stereo cameras due to environmental differences.
For a technique to sufficiently generalize across many domains and leverage all of the data available for best performance, it must support both of these configurations.

Based on these baseline requirements, three stand out: ORB-SLAM3, OpenVSLAM, and RTABMap. 
Some attention should be paid to VINS-Fusion, which met all of these requirements - and more - except for support for RGB-D sensors. 
While it would have been possible to compare this technique with the limited sensor support that it contains, it has already been shown to under-perform ORB-SLAM3 in several domains of interest in this study \cite{ORB-SLAM3}.
Thus, this method is not chosen for comparison.

SVO, DSO, and OKVIS are all popular Visual or Visual-Inertial Odometry methods and were not considered for comparison due to their lack of loop closure and RGB-D support.
These VO methods cannot be considered an alternatives for service robots due to the lack of globally consistent representations often required to perform application tasks.
Especially for indoor and long-running service applications, having accurate measures of distance and poses are important to safe high-speed navigation.
These are known to be high-quality and efficient VO methods but they are not suitable for service robot deployment to replace existing localization methods.

LSD-SLAM is capable of operating in large environments via direct alignment of images leveraging intensity gradients.
It can provide reconstructions of semi-dense probabilistic depth maps \cite{LSD-SLAM1}.

While it supports re-localizing, loop closures, and stereo cameras (in addition to monocular cameras), previous experiments have shown LSD-SLAM achieves lower accuracy than the ORB-SLAM family \cite{ORB-SLAM2}.
Further, it does not support RGB-D sensors and pure localization on previously stored map, two required features for a significant proportion service robot applications.

PL-SLAM is a stereo SLAM which utilizes point and line segment features.
This approach is essential for environments with low texture.
In these situations, traditional VSLAM approaches perform poorly compared to PL-SLAM \cite{PL-SLAM}.
However, due to the absence of a localization mode and numerous run-time failures of the algorithm, it is unsuitable for comparison.

Kimera is a promising Visual-Inertial Odometry (VIO) algorithm.
It has such techniques as PGO and loop closure detection, 3D mesh reconstruction and semantic labeling already developed \cite{Kimera}.
While this method is young and still being actively developed, is has shown great potential to become a basis for a mature, robust, and feature complete VSLAM.
During initial testing, this algorithm was selected for quantitative assessment.
However, Kimera was shown to be significantly unstable resulting in frequent crashes and exits while executing on several datasets.
Thusly, this algorithm, while feature rich, is not yet suitable for further analysis.

\section{Visual SLAM Approaches}
\label{sec:implementation}

\subsection{ORB-SLAM3}
ORB-SLAM3 is the continuation of the ORB-SLAM project: a versatile visual SLAM sharpened to operate with a wide variety of sensors (monocular, stereo, RGB-D cameras).
It uses the ORB feature to provide short and medium term tracking and DBoW2 for long term data association.
The ORB-SLAM family of methods has been a popular mainstay since 2015 and has been adapted numerous times to augment odometry systems for service robots. 
This technique has contained re-localization, pure localization, and loop-closure capabilities since its first release.
The third version of ORB-SLAM has introduced some important features over its predecessors \cite{ORB-SLAM3}:
\begin{itemize}
    \item Visual-Inertial SLAM - Sensor fusion of visual and inertial sensors resulting in more robust tracking in situations having few point features (e.g. in corridors, low-texture environments, motion blur caused by fast movements)
    \item Multi-map - Continuing a previous SLAM session
    \item Fisheye camera support - Widening platform support to other common camera types
\end{itemize}

\subsection{RTABMap}
RTABMap is the one of the oldest mixed-modality SLAM approaches.
It was first released in 2013 and still is under active maintenance and support \cite{RTABMap}.
It covers a wide variety of input sensors including not only stereo, RGB-D and fisheye cameras, but also odometry and 2D/3D lidar data.
This makes RTABMap a flexible SLAM approach unique from the other methods studied.
RTABMap has long been a fixture of the Robot Operating System (ROS) as an alternative to 2D SLAM, sometimes used in concert with mobile robot navigation.
Rather than creating feature maps, RTABMap creates dense 3D and 2D representation of the environment that can be used analogous to a pure-2D lidar SLAM.
This allowed it to be a drop-in semi-visual SLAM replacement to existing methods without any further post-processing - a unique and compelling advantage.
RTABMap is already in use in hobby, research, and small scale service robot applications today as a 2D SLAM "alternative" (often paired still with lidar or depth data).
Unsurprisingly, it contains all of the basic features for utilization in mobile robot applications.

\subsection{OpenVSLAM}
OpenVSLAM is a well engineered, module-structured implementation of an ORB feature based Visual graph SLAM \cite{OpenVSLAM}.
It contains optimized implementations of feature extractors and stereo matchers.
Additionally, the authors have developed a unique frame tracking module allowing it to perform fast and accurate localization, competitive with much of the state of the art.
It is also able to leverage many more cameras models than a typical VSLAM implementation, notably including equirectangular and fisheye.
OpenVSLAM was also developed for scalability and integration into practical applications rather than solely for research novelty.
Licensed under BSD-2.0, it is designed for expansion and has seen adoption by both research and industrial users that have previously shied away from other VSLAM library's restricted licenses.
OpenVSLAM contains all of the major features of interest of the ORB-SLAM family (in some cases, more) except support for IMU fusion.
This method has advantages in that it also has an active community developing and maintaining it and was well designed by experienced engineers to be more easily understood, tested, and expanded over time.
Compared to both RTABMap and ORB-SLAM3, OpenVSLAM does not have the same history of mobile and service robot use for odometry or mapping.
However, OpenVSLAM is a far newer approach (2019) compared to ORB-SLAM (2015) and RTABMap (2013); the authors do not feel this is due to any technical inferiority.

\section{System Setup}
\label{sec:description}

\subsection{Target Service Platforms}
\label{sec:descriptionA}

Service robots are deployed to accomplish a broad range of applications whose bounds are often ill-defined.
From warehouse robots, to food delivery platforms, to surveillance drones, there is a broad range of both types of hardware and environments which may be considered "service robot" related.
This work seeks to identify general purpose VSLAM techniques that can reliably operate in this broad range of applications with various sensors.
The following list contains the primary attributes of service robots for which this work is conducted in accordance with.

\begin{itemize}
    \item Wheel, legged, or other ground-based platforms or Micro Aerial Vehicles (MAV) performing commercial or consumer assistive functionality
    \item The platform should be supplied with a commonly available passive stereo or active RGB-D sensor of sufficient range for its environment
    \item The platform may, but is not required to, provide a camera with synchronized IMU for inertial fusion
    \item The robot should contain at least the minimum compute resources on-board to run navigation, sensor-based collision avoidance, and VSLAM pipelines (e.g. 5th Generation Intel Core i5 or newer, Nvidia Jetson series, etc)
    \item The operating environment of the robot is not restricted and may include indoor, outdoor urban, and outdoor unstructured areas - or any combination thereof.
\end{itemize}

\subsection{Environment for Experiments}

For these experiments, a PC powered by an Intel Core i5-6600 4-core CPU operating at 3.30GHz with 8 GB of RAM memory was utilized.
This 6th generation configuration is similar to those found on many service and industrial mobile platforms.
While Intel's newest line of processors are the 11th generation, many equipment manufacturers have lagged behind the newest in consumer electronics to reduce costs and integrate processors into fan-less configurations.
64-bit Ubuntu 18.04 was used for all of the experiments.

The experiments used the latest versions available at the time of publication (Feb 2021) of each visual SLAM technique.
All algorithms were build with "Release" flag and options recommended by each method.

\begin{table*}[ht] 
 \caption{Datasets comparison}
 \label{tab:datasets}
 \begin{center}
 \begin{tabular}{ |c|c|c|c|c|c|c|c|c| }
   \hline
   \multirow{2}{*}{Dataset} & \multicolumn{3}{c|}{Sensors} & \multirow{2}{*}{GT} & \multicolumn{3}{c|}{Environment} & \multirow{2}{*}{Platform} \\
   \cline{2-4}\cline{6-8}
   & Stereo & RGB-D & IMU & & Small Indoor & Large Indoor & Outdoor & \\
   \hline
   \hline
   EuRoC MAV & \checkmark & & \checkmark & \checkmark & \checkmark & \checkmark & & Drone \\
   TUM RGB-D & & \checkmark & & \checkmark & \checkmark & \checkmark & & Robot, Hand \\
   KITTI & \checkmark & & & \checkmark$^1$ & & & \checkmark & Car \\
   \hline
   \multicolumn{9}{l}{} \\
   \multicolumn{9}{l}{$^1$ Ground Truth for the trajectories available only for first 11 sequences} \\
 \end{tabular}
 \end{center}
\end{table*}

\section{Datasets}
\label{sec:datasets}

To estimate the performance of these approaches, datasets are required representing a variety of different environments and sensor inputs that span the requirements in Section \ref{sec:descriptionA}.
While all environments service robots may be deployed in are of interest, in practice, there are a few classifications of spaces that are most common for service robots.
It is useful to classify these common operational environments into three distinct groups for focused study and analysis.
These three environment types are used as surrogates to compare the viability of the techniques:

\begin{itemize}
 \item Small Indoors - Small human structures (houses, offices)
 \item Large Indoors - Large but bounded spaces (warehouses)
 \item Urban Outdoors - Open-air, urban city and suburban areas
\end{itemize}

For all of these environments, it is common to find ground-based robots.
However, this study should not be limited only to ground-based robots but also include MAVs (coinciding with the popularity of drones).
At least one of these environmental datasets should also contain data collected from a MAV such that this comparison is representative of the definition of a service robot set forth in Section \ref{sec:descriptionA}.

Further to this point, it is important to also consider the sensors being utilized for comparison.
VSLAM on robotic platforms mostly relies on cameras (monocular, stereo or RGB-D); it is important datasets span this range of inputs, at minimum.
However, some modern cameras marketed towards service robots also include hardware-synchronized IMUs, enabling advanced sensor fusion techniques \cite{d435}.
Not all robots will include these types of sensors, but enough may that it is worth evaluating our visual SLAM techniques on their merits both with and without IMU fusion, if available.

The Table \ref{tab:datasets} shows the popular datasets used for this benchmark and how they meet these conditions.
While none show explicit monocular data collection, obviously any stereo or RGB-D configuration can provide monocular support as well.
These datasets were carefully chosen in accordance with the discussion above to contain a fair mixture of environment, sensor support, and robot type to be as closely representative of general purpose service robot applications as possible.

\subsection{EuRoC MAV}
\begin{figure}[ht]
  \centering
  \setlength{\belowcaptionskip}{-2pt}
  \includegraphics[scale=0.14]{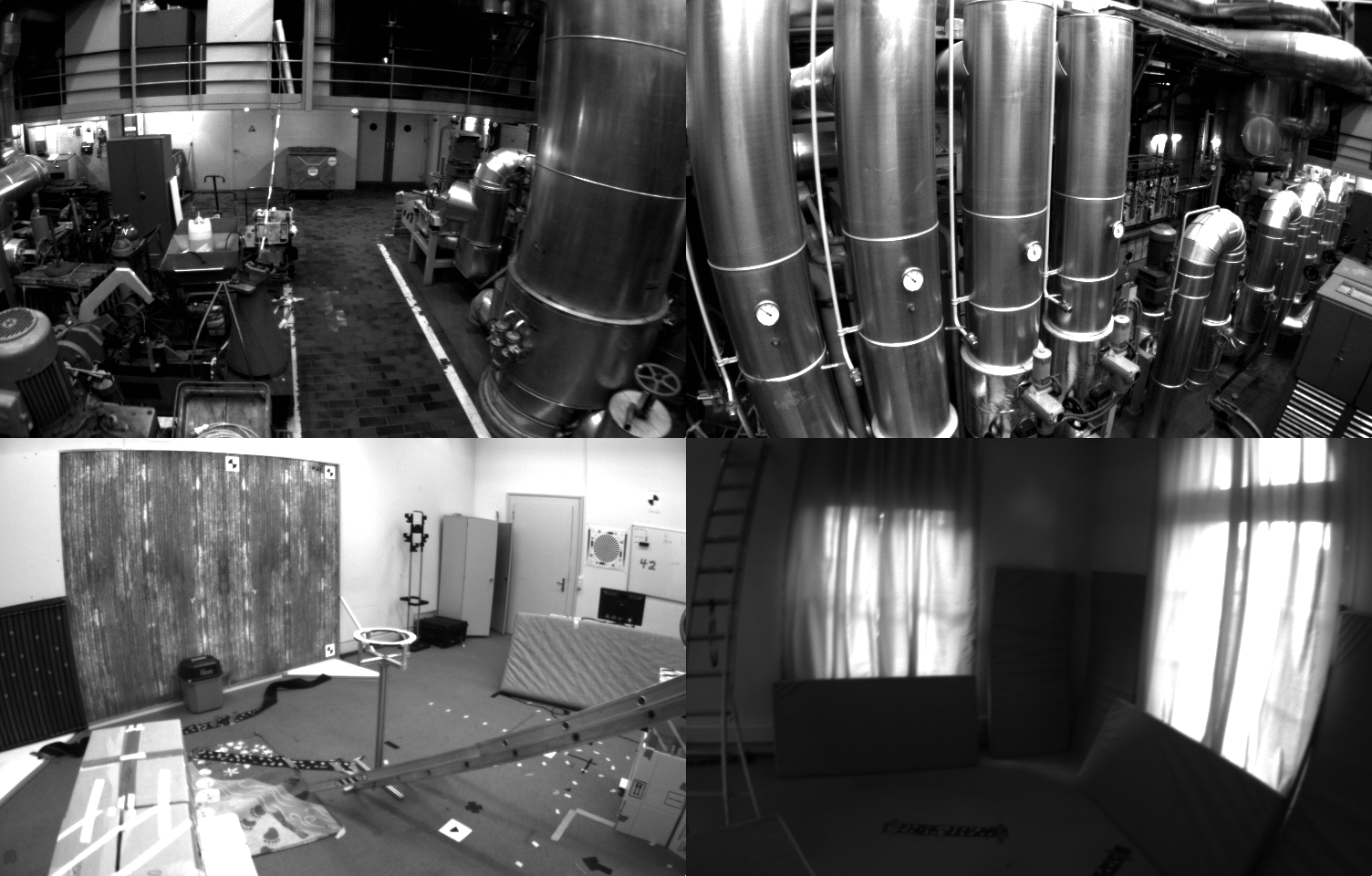}
  \caption{EuRoC scenes example: Large indoor facility (top row) and small office (bottom row) \cite{EuRoC}}
  \label{fig:EuRoC}
\end{figure}

Among the most popular and used in literature benchmarks, EuRoC MAV represents visual-inertial (VI) sequences collected on-board a MAV \cite{EuRoC}.
The sequences are divided into small indoor office and large indoor factory chains.
Each sequence is ranked by its complexity expressed in drone travel and rotation speed, scene change rate, the presence in the camera frame variations of illumination.
An example of scenes for this dataset is shown in Figure \ref{fig:EuRoC}.
Having a stereo camera and IMU on-board, this dataset allows for visual SLAM estimation in a variety of indoor environments covering monocular and stereo modes, with or without using IMU sensor fusion, if available.

\subsection{TUM RGB-D}
\begin{figure}[ht]
  \centering
  \setlength{\belowcaptionskip}{-2pt}
  \includegraphics[scale=0.17]{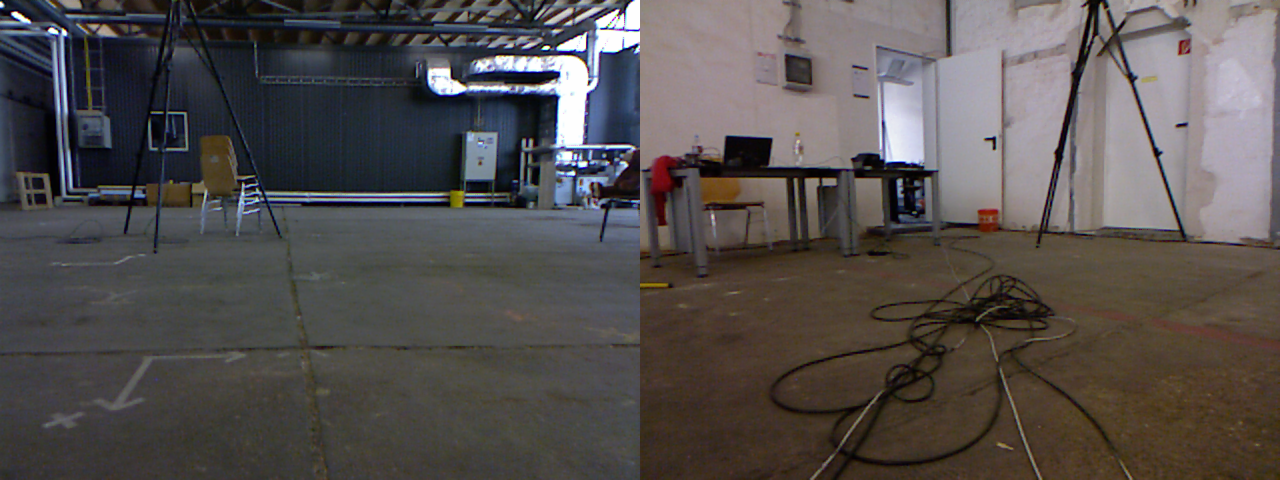}
  \caption{TUM RGB-D scenes example for Pioneer robot-based sequences \cite{TUM-RGBD}}
  \label{fig:TUM-RGBD}
\end{figure}

The TUM RGB-D is a dataset containing color and depth cameras \cite{TUM-RGBD}.
Sequences in the dataset are divided into: handheld SLAM sequences, ground-based robot SLAM sequences, and sequences that contain dynamic objects or targeted at 3D reconstruction.
However, for service robots, the most interesting sequences are those of the "Pioneer" robot SLAM sessions.
This is a wheel-based platform with a Kinect camera mounted on top.
In this group, there are 4 sequences recorded in an indoor garage with a low texture environment.
An example of scenes for this dataset is shown in a Figure \ref{fig:TUM-RGBD}.
This dataset uniquely offers an indoor ground-based robot perspective with a target sensor and has not yet been benchmarked with ORB-SLAM3 and OpenVSLAM.

\subsection{KITTI}
\begin{figure}[ht]
  \centering
  \includegraphics[scale=0.17]{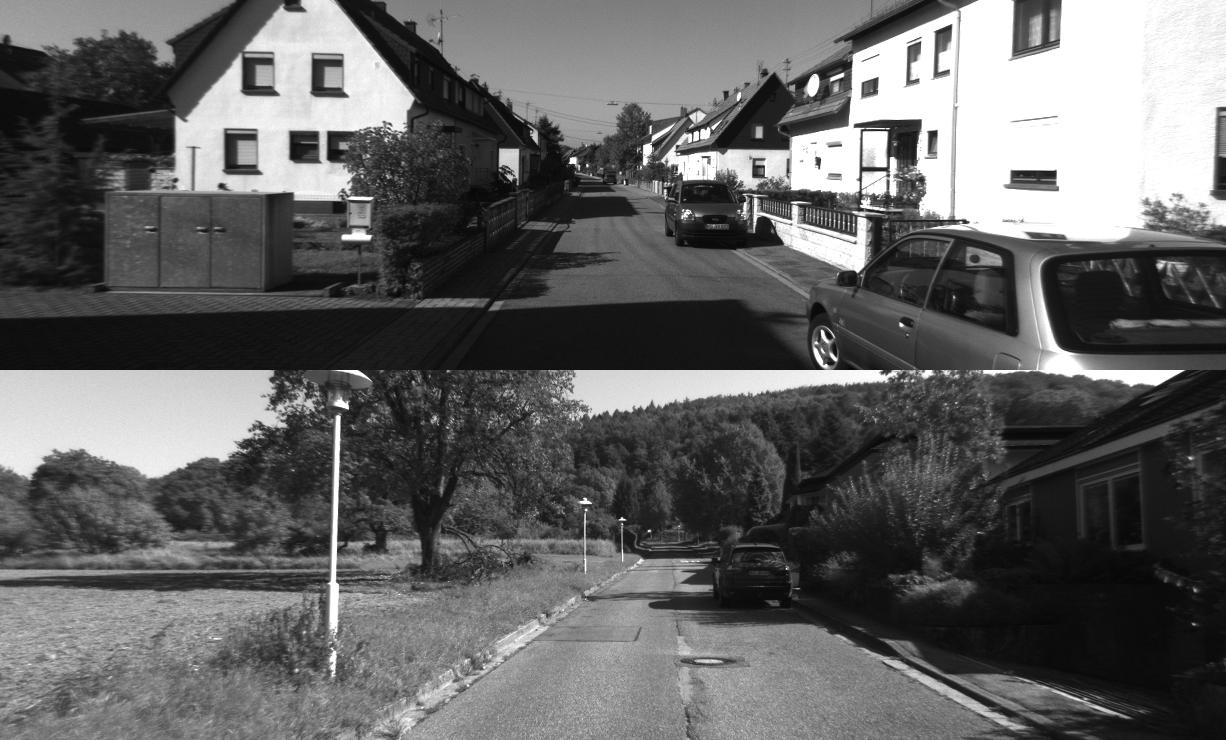}
  \caption{KITTI scenes example: city streets (top image) and suburban areas (bottom image) \cite{KITTI}}
  \label{fig:KITTI}
\end{figure}

For outdoor urban environments, the KITTI dataset was used for comparison.
This dataset was recorded by using a vehicle outfitted with modern autonomous driving sensors, including a passive stereo camera on top \cite{KITTI}.
It contains 22 sequences but only the first 11 have ground truth information for a formal comparison.
An example scene is shown in a Figure \ref{fig:KITTI}.

\section{Experiments and Analysis}
\label{sec:experiments}

The evaluation of general purpose visual SLAM techniques was divided into two categories: the evaluation of indoor SLAM measured on EuRoC and TUM RGB-D and the evaluation of outdoor SLAM measured on KITTI.
This organization is made to compare the techniques in each fundamental service robot domain separately and then study the overall performance.
The algorithms are tested on the three datasets in the modes described below:
\begin{itemize}
 \item EuRoC: ORB-SLAM3 w/ monocular, stereo, monocular-inertial, and stereo-inertial; OpenVSLAM w/ monocular, stereo; RTABMap w/ stereo
 \item TUM RGB-D: All methods w/ RGB-D
 \item KITTI: ORB-SLAM3 w/ monocular and stereo; OpenVSLAM w/ monocular and stereo; RTABMap w/ stereo
\end{itemize}

Note: ORB-SLAM3 is the only method that supports inertial fusion and RTABMap does not support monocular SLAM.

We utilized the tuned parameters provided by the authors for each algorithm such that the outputs were consistent with the previously recorded results.
For evaluating the accuracy in each experiment, the poses were gathered directly after execution. This allowed only for pose optimizations produced live during the experiment without any post-processing.

Estimated RMSE ATE is the representative measure for SLAM accuracy in these experiments \cite{RMS-ATE}.
Calculation of RMSE was made by comparing the VSLAM trajectory with the ground truth information provided.
In monocular sequences, the trajectory poses are not properly scaled due to the impossibility of precise distance measurement by means of using only a single passive camera.
Therefore, alignment in monocular cases should use a Sim(3) transformation.
For stereo and RGB-D cases, where the calculation of the robot trajectory should provide accurate scale, the SE(3) transformation is utilized.
For visual-inertial cases, ORB-SLAM3 can recover the scale, roll, and pitch angles, thus the alignment was made using a 4DoF transformation.
In all cases, the closed-form solution of the Horn algorithm was used for trajectories alignment \cite{Horn}.

The results in Tables \ref{tab:results-euroc},  \ref{tab:results-tum},  \ref{tab:results-kitti} below represent the mean RMSE over 10 executions of each sequence.
Additionally, VSLAM stability is partially estimated using the standard deviation of RMSE.
Stability is also measured in crashing, early termination, lose of track, etc.
Each table also contains runtime duration of one dataset execution in mm:ss format.

\subsection{Indoor SLAM}

\begin{table*}[ht]
 \caption{EuRoC MAV benchmarking results}
 \label{tab:results-euroc}
 \begin{center}
 \begin{tabular}{ |c|c|c|c|c|c|c|c|c|c|c|c|c|c| }
   \hline
   & & MH01 & MH02 & MH03 & MH04 & MH05 & V101 & V102 & V103 & V201 & V202 & V203 & Avg / Dur \\
   \hline
   \hline
   ORB-SLAM3 & RMSE,m &0.0276&0.0546&\textbf{0.0307}&\textbf{0.139}&0.2134&\textbf{0.0332}&\textbf{0.139}&0.330&\textbf{0.0213}&0.070&0.190&0.113 \\
   (mono) & st.dev,m &0.0210&0.0591&0.0039&0.106&0.4265&0.0004&0.243&0.379&0.0059&0.142&0.254&16:27 \\  
   \hline
   ORB-SLAM3 & RMSE,m &0.0262&0.0680&0.0315&0.149&0.0908&0.0441&\textbf{\textit{0.018}}&\textbf{\textit{0.025}}&0.0465&\textbf{\textit{0.023}}&\textbf{\textit{0.026}}&\textbf{\textit{0.050}} \\
   (monoi) & st.dev,m &0.0053&0.0191&0.0010&0.023&0.0321&0.0027&0.002&0.004&0.0045&0.004&0.006&18:23 \\
   \hline
   OpenVSLAM & RMSE,m &\textbf{0.0198}&\textbf{0.0187}&0.0367&0.186&\textbf{0.0588}&0.0497&0.301&\textbf{0.054}&0.0265&\textbf{0.027}&\textbf{0.114}&\textbf{0.081} \\
   (mono) & st.dev,m &0.0031&0.0027&0.0049&0.033&0.0126&0.0408&0.552&0.029&0.0083&0.010&0.054&12:44 \\
   \hline
   \hline
   ORB-SLAM3 & RMSE,m &0.0364&\textbf{0.0193}&\textbf{0.0265}&0.114&0.0563&\textbf{0.0347}&\textbf{0.022}&0.051&0.0427&\textbf{0.027}&0.528&0.087 \\
   (stereo) & st.dev,m &0.0029&0.0021&0.0026&0.023&0.0126&0.0011&0.001&0.016&0.0089&0.005&0.279&23:27 \\
   \hline
   ORB-SLAM3 & RMSE,m &0.0339&0.0410&0.0295&\textbf{\textit{0.055}}&0.0988&0.0367&\textbf{\textit{0.014}}&\textbf{\textit{0.024}}&\textbf{\textit{0.0382}}&\textbf{\textit{0.019}}&\textbf{\textit{0.040}}&\textbf{\textit{0.039}} \\
   (stereoi) & st.dev,m &0.0041&0.0066&0.0018&0.003&0.0109&0.0007&0.002&0.001&0.0028&0.002&0.014&26:14 \\
   \hline
   OpenVSLAM & RMSE,m &0.0301&0.0239&0.0323&\textbf{0.082}&\textbf{0.0557}&0.0358&0.034&\textbf{0.034}&\textbf{0.0425}&0.038&\textbf{0.169}&\textbf{0.052} \\
   (stereo) & st.dev,m &0.0033&0.0030&0.0028&0.015&0.0046&0.0004&0.045&0.011&0.0253&0.010&0.039&21:04 \\
   \hline
   RTABMap & RMSE,m &\textbf{0.0280}&0.0374&0.0691&0.429&0.1890&0.1142&0.112&0.238&0.1185&0.512&-&0.185 \\
   (stereo) & st.dev,m &0.0000&0.0000&0.0000&0.000&0.0000&0.0000&0.000&0.000&0.0000&0.000&-&18:39 \\
   \hline
 \end{tabular}
 \end{center}
\end{table*}

Table \ref{tab:results-euroc} shows the results of ORB-SLAM3, OpenVSLAM and RTABMap benchmarked on EuRoC MAV dataset, grouped by monocular and stereo runs.
In each group, the best results are marked in a \textbf{bold} font.
However, it is necessary to study both inertial fusion and non-inertial techniques separately due to our service robot definition set forth in Section \ref{sec:descriptionA}.
In the case where a VI method outperforms a pure visual method, the VI method is marked in \textbf{\textit{bold+italics}} while the best non-inertial option is simply in \textbf{bold}.

All of the techniques in non-inertial monocular mode were unstable, showing a comparatively large variation from run to run on the EuRoC dataset.
In several sequences, both monocular supported algorithms failed to determine sensor poses causing enormous drift throughout the trajectories.
For ORB-SLAM3 this effect was noted on 7/11 sequences, while for OpenVSLAM, only 4/11 sequences.
The ORB-SLAM3 inertial fusion technique drastically reduced this behavior for nearly all monocular sequences.
This significant performance boost points to ORB-SLAM3 with inertial fusion as clearly the best monocular technique.
However, while monocular vision is a compelling area of research for the future, service robots are nearly always outfitted with stereo or RGB-D sensors for collision avoidance.
Thusly, all of the monocular approaches are of questionable benefit when additional information is available.

Using EuRoC stereo information, OpenVSLAM and both modalities of ORB-SLAM3 performed well.
It is useful to compare both visual-only and separately all stereo results for the different common types of sensors service robots may be built with.
While some stereo sensors contain hardware-synchronized IMUs, many do not and this general purpose analysis should cover the range of hardware commonly employed by service robots.
Note that RTABMap's results were stable throughout this experiment, albeit some re-localizations.
However, it performed overall much worse on all but one EuRoC MAV sequence compared with the other algorithms and is evidently less well suited for 6DOF stereo use.

When comparing the remaining non-inertial-fusion stereo techniques on the EuRoC dataset, OpenVSLAM's overall performance is prominent ($\sim5cm$ accuracy).
In most sequences, stereo ORB-SLAM3 and OpenVSLAM had somewhat similar performance - with the notable exception of \texttt{V203}.
\texttt{V203} is a fast moving sequence with motion blur for which RTABMap could not produce meaningful tracks. 
OpenVSLAM and ORB-SLAM3 without IMU fusion both displayed lower than nominal positioning accuracy in \texttt{V203} and frequently had to re-localize due to lost tracking.
However, OpenVSLAM was able to far more reliably keep its track and consistently keep its positioning accuracy under $20 cm$ (3x better than ORB-SLAM3).
On the same sequence, ORB-SLAM3 completely floundered and was unable to reliably maintain a track and the standard deviation was $0.254 m$ across the 10 executions.
In other complex sequences, such as \texttt{MH04} and \texttt{V103}, OpenVSLAM also outperformed pure stereo ORB-SLAM3 in low-lighting and rapidly changing lighting conditions.

\begin{figure}[ht]
  \includegraphics[scale=0.131]{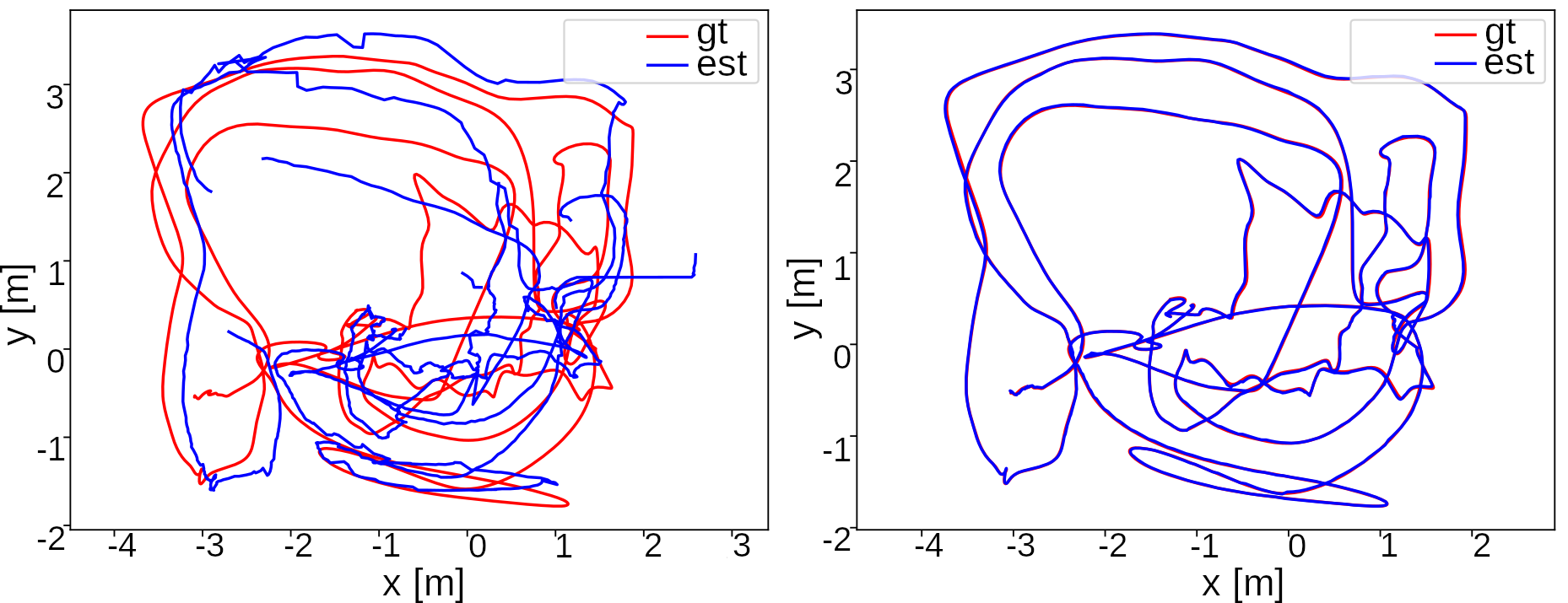}
  \caption{Pure stereo mode (left) and IMU sensor fusion effect on stereo SLAM (right) for ORB-SLAM3 on EuRoC V203 sequence. Each image on the figure contains an estimated trajectory (est) drawn over ground truth (gt).}
  \label{fig:ORB3-stereo-VI}
\end{figure}

When comparing all stereo results, the inertial fusion employed by ORB-SLAM3 is greatly beneficial.
It further aids in avoiding lost tracks in particularly difficult cases, including \texttt{V203}.
Figure \ref{fig:ORB3-stereo-VI} illustrates visually the result of inertial fusion in ORB-SLAM3 on \texttt{V203}.
The left image has the track from ORB-SLAM3 using stereo while right has the track with stereo-inertial fusion.
The VI mode in ORB-SLAM3 allowed it to continue tracking when visual-only tracking was clearly lost and was unable to fully re-localize.
While this was the only sequence of EuRoC to exhibit significant differences between the stereo modes, it does highlight an edge case that the VI mode fills especially well.

From the stereo results on EuRoC, it is clear that OpenVSLAM is a leader in non-inertial fusion while ORB-SLAM3's VI mode is the overall best performing.
As this analysis seeks to support general purpose service robots, it is worth noting that the performance of stereo OpenVSLAM outperforms in sequences with variable or poor lighting conditions. 
In the large scale sequences recorded in a factory setting, both approaches produced very similar results.
Either option provides a suitable solution for service robots equipped with stereo cameras indoors.

\begin{table*}[ht]
 \caption{TUM RGB-D benchmarking results}
 \label{tab:results-tum}
 \begin{center}
 \begin{tabular}{ |c|c|c|c|c|c|c|c| }
   \hline
   & & 360 & slam & slam2 & slam3 & Avg & Dur \\
   \hline
   \hline
   \multirow{2}{*}{ORB-SLAM3 (RGB-D)} & RMSE,m &1.059&2.118&1.709&1.992&1.720&03:35 \\
   & st.dev,m &0.529&0.189&0.188&0.119&& \\
   \hline
   \multirow{2}{*}{OpenVSLAM (RGB-D)} & RMSE,m &0.048&\textbf{0.188}&1.224&\textbf{0.106}&0.392&03:26 \\
   & st.dev,m &0.010&0.003&0.011&0.013&& \\
   \hline
   \multirow{2}{*}{RTABMap (RGB-D)} & RMSE,m &\textbf{0.038}&0.351&\textbf{0.044}&0.716&\textbf{0.287}&05:20 \\
   & st.dev,m &0.000&0.000&0.000&0.000&& \\
   \hline
 \end{tabular}
 \end{center}
\end{table*}

Table \ref{tab:results-tum} presents the results for ground-robot experiments in a low-featured indoor facility measured using the TUM RGB-D dataset's four Pioneer sequences.
All of the methods showed low accuracy - frequently losing tracks and incorrectly re-localizing.
Since this robot experienced much slower rotations and motion compared with the EuRoC drone experiments, the root cause of this behavior is the low-texture environment in all sequences.
These conditions did not allow for the discovery of sufficient point features to localize reliably in any approach.
This makes a particularly valuable comparison to analyze how well these techniques perform in a degraded setting.

RTABMap is the leading method in this environment; it is $36\%$ more accurate than OpenVSLAM and over $499\%$ compared to ORB-SLAM3. 
Through the sequences, RTABMap shown the best ability to recover from loosing its track.
However, under these conditions, its accuracy was still wildly variable between $3.8 cm$ to over $71 cm$.
This performance isn't considered exceptional, especially in \textit{pioneer\_slam} and  \textit{pioneer\_slam3}, but it does show reasonable re-localization capabilities even in poor lighting conditions.

\begin{figure}[ht]
  \centering
  \includegraphics[scale=0.115]{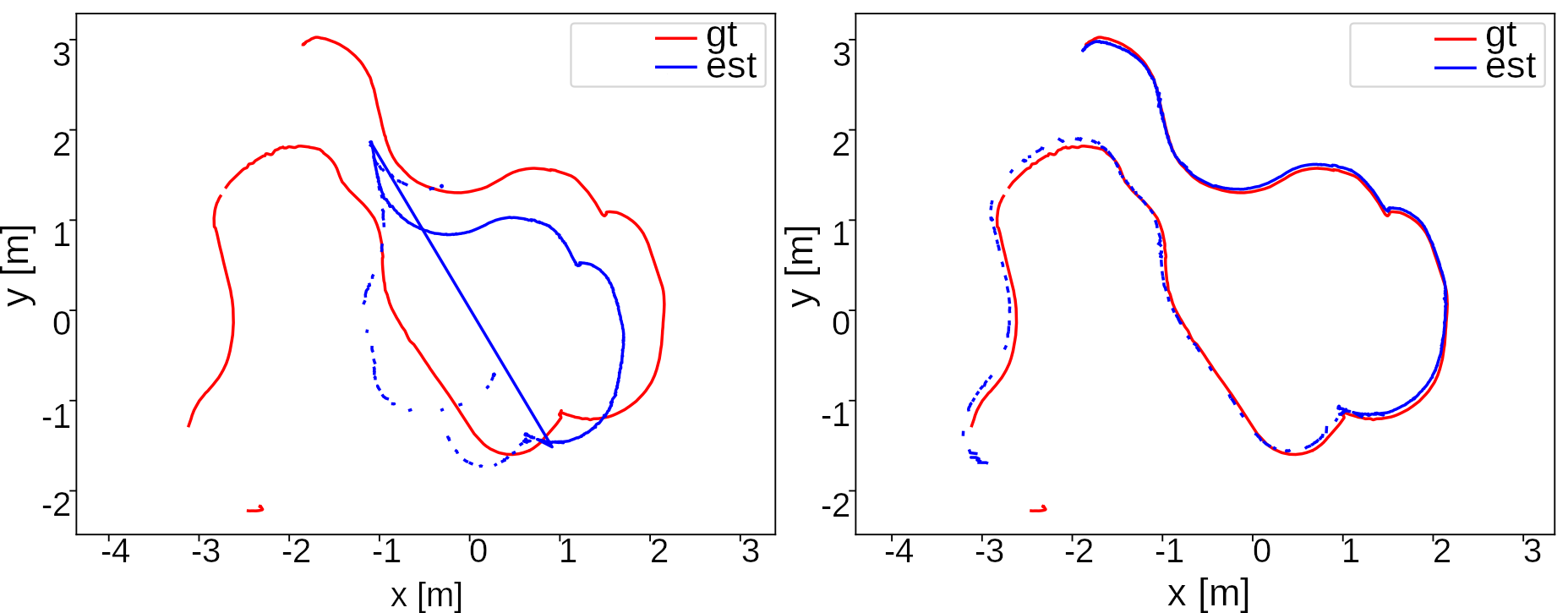}
  \caption{ORB-SLAM3 (left) and OpenVSLAM (right) on TUM RGB-D pioneer\_slam3 sequence. Each image on the figure contains estimated trajectory (est) drawn over ground truth (gt).}
  \label{fig:TUM_slam3}
\end{figure}

While OpenVSLAM is behind in overall accuracy, it did perform comparatively well in 3 of the 4 sequences.
OpenVSLAM had issues maintaining localization in \textit{pioneer\_slam2} but it was the only technique which otherwise maintained a less than $20 cm$ accuracy with low variance.
While it also lost its track on occasion, OpenVSLAM was exceptionally good at re-localizing reliably, even in degraded conditions.
OpenVSLAM and RTABMap outperformed ORB-SLAM3 in all sequences of TUM RGB-D.
Incorrect re-localization of ORB-SLAM3 on most sequences led to incorrect map stitching and the average RMSE was never less than $1.0$ meter, shown in Figure \ref{fig:TUM_slam3}.
The high standard deviation also points to an irregular ability to recover from a lost track in these non-optimal conditions.
From these results, RTABMap and OpenVSLAM are the two most reasonably performing methods in low-texture, ground-based service robot conditions (while ORB-SLAM3 performed particularly poorly).

\subsection{Outdoor SLAM}
\begin{table*}[!ht]
 \caption{KITTI benchmarking results}
 \label{tab:results-kitti}
 \begin{center}
 \begin{tabular}{ |c|c|c|c|c|c|c|c|c|c|c|c|c|c| }
   \hline
   & & 00 & 02 & 03 & 04 & 05 & 06 & 07 & 08 & 09 & 10 & Avg & Dur \\
   \hline
   \hline
   \multirow{2}{*}{ORB-SLAM3 (mono)} & RMSE,m &9.109&31.928&2.503&1.281&6.351&\textbf{15.149}&\textbf{3.931}&71.197&\textbf{17.452}&9.883&\textbf{16.878}&15:04 \\
   & st.dev,m &1.175&17.568&0.626&0.303&1.117&0.660&0.246&54.604&11.157&1.514&& \\
   \hline
   \multirow{2}{*}{OpenVSLAM (mono)} & RMSE,m &\textbf{6.860}&\textbf{31.434}&\textbf{1.632}&\textbf{0.372}&\textbf{5.897}&18.215&5.063&\textbf{70.470}&29.923&\textbf{9.454}&17.932&15:12 \\
   & st.dev,m &1.188&8.035&0.573&0.147&0.702&1.644&0.459&4.814&14.898&1.458&& \\
   \hline
   \hline
   \multirow{2}{*}{ORB-SLAM3 (stereo)} & RMSE,m &1.348&7.114&0.759&0.249&\textbf{0.778}&\textbf{0.783}&0.554&3.619&\textbf{1.640}&\textbf{1.030}&1.787&28:27 \\
   & st.dev,m &0.035&0.619&0.053&0.043&0.045&0.107&0.043&0.185&0.045&0.101&& \\
   \hline
   \multirow{2}{*}{OpenVSLAM (stereo)} & RMSE,m &\textbf{1.339}&\textbf{5.695}&\textbf{0.725}&\textbf{0.197}&0.783&0.899&\textbf{0.547}&\textbf{3.446}&2.175&1.235&\textbf{1.704}&30:25 \\
   & st.dev,m &0.020&0.371&0.059&0.037&0.016&0.090&0.057&0.124&0.611&0.045&& \\
   \hline
   \multirow{2}{*}{RTABMap (stereo)} & RMSE,m &1.370&5.824&1.694&1.389&1.715&2.278&0.665&7.428&4.683&1.120&2.817&15:12 \\
   & st.dev,m &0.000&0.000&0.000&0.000&0.000&0.000&0.000&0.000&0.000&0.000&& \\
   \hline
 \end{tabular}
 \end{center}
\end{table*}

Table \ref{tab:results-kitti} shows the results from the KITTI dataset, with \textbf{bold} as the best results and grouped similar to Table \ref{tab:results-euroc}.
The \texttt{KITTI\_01} sequence was excluded from analysis due to its instability and unrepresentative nature for this service robot-centric comparison.
ORB-SLAM3 and OpenVSLAM failed to produce results and were unable to gather the required features for stable tracking.
\texttt{KITTI\_01} is a fast rural highway sequence with few unique features; operating at speeds and in an environment incompatable with service robots.
Thusly, this sequence was not considered in our analysis.

Monocular sequences resulted in inaccurate trajectories for outdoor applications, very similarly to the indoor cases.
Both monocular algorithms failed to calculate pose changes which caused increasing drift throughout the trajectories.
This effect was noted on ORB-SLAM3 and OpenVSLAM in 4/10 and 3/10 sequences, respectively.
The positioning accuracy of both algorithms is deeply insufficient for any practical application, with over $16m$ of RMSE drift each, on average. 
Hence, any of these monocular VSLAM's for outdoor use is not recommended.

The results improve markedly in stereo configurations.
From the stereo results outdoors, ORB-SLAM3 and OpenSLAM showed very similar performance, at around $1.7 m$ average accuracy for outdoor environments, where OpenVSLAM slightly outperformed ORB-SLAM3 ($< 5\%$).
It is worth noting that ORB-SLAM3 showed instability causing crashes from segmentation faults.
During the experiments, these crashes were noted on \texttt{KITTI\_04} and \texttt{KITTI\_06} sequences especially.
RTABMap showed very stable but the least accurate results on nearly all sequences.
It contained a perceptible pose drift on many sequences where other methods were very reliable.

OpenVSLAM is only \textit{slightly} more accurate than ORB-SLAM3.
However, ORB-SLAM3's crashes make it considerably less reliable; making it unsuitable for long-duration use.
This fact, along with a slight improvement in accuracy, causes OpenVSLAM to be considered the single most suitable method for urban outdoor use from our analysis.

\section{Limitations}
\label{sec:limitations}
As the experiments have shown, despite the great performance of visual SLAM techniques in ideal conditions, scenes with low or frequently changing lighting conditions and fast movements cause significant drops in re-localization capabilities.
Any service robot application making use of VSLAM must still consider the lighting conditions of their environment in assessing whether this technology is adequate for their needs.
Re-localization and tracking in variable lighting conditions is still a limitation of all studied visual SLAM methods.

This analysis did not study the long-term pure localization capabilities of these visual SLAM techniques.
Service robot environments are almost always changing and old feature sets recorded may become outdated over time.
It is left as future work to analyze the technique's long term viability in a single environment as the environment changes and study how it degrades localization performance.

\section{Conclusion}
\label{sec:conclusion}

The comparison of 3 feature-rich and mature visual SLAM approaches has been provided across many operating domains of service robots.
The adoption of VSLAM approaches on service robots is a critical evolution to grow the industry away from prohibitively expensive lidars.
Service robots are deployed in many different environments, with any different sensors, on the ground and in the air.

It was determined that for indoor environments with stereo sensors, OpenVSLAM and ORB-SLAM3 with inertial fusion were the best performing.
With RGB-D sensors, OpenVSLAM and RTABMap performed well, even in degraded low-feature conditions.
Outdoors, OpenVSLAM and ORB-SLAM3 shined with remarkably similar performance.
However, ORB-SLAM3 displayed comparatively poorer reliability in several experiments.

The goal of this analysis is to identify general purpose techniques which may be used to support innumerable service robot applications.
Through this experimentation, it was concluded that OpenVSLAM was the overall best general purpose technique for the broadest range of service robot types, environments, and sensors.
It performed well in all three studies, showcasing superior re-localization, variable lighting performance, with high reliability.
However, when a synchronized stereo-IMU pair is available in good lighting conditions, ORB-SLAM3 with inertial fusion is another suitable option - but ORB-SLAM3 is not recommended due to periodic segmentation faults.
While RTABMap has shown lower overall performance, it was the most deterministic in evaluations, containing no variation in all 250 executions.
Finally, due to improvements of overall accuracy and robustness of performance, incorporating an IMU fusion technique is typically preferred when it possible.

The results of this evaluation will further be used to impact and accelerate the community's use of VSLAM for robotics.
OpenVSLAM has been integrated into the ROS 2 Nav2 System (a.k.a. ROS 2 Navigation) as an alternative SLAM and localization system parallel to 2D SLAM Toolbox and AMCL localization \cite{Nav2, SlamToolbox, SlamToolbox2, AMCL}.

\end{document}